\documentclass[10pt,twocolumn,letterpaper]{article}

\usepackage{iccv}
\usepackage{times}
\usepackage{epsfig}
\usepackage{graphicx}
\usepackage{amsmath}
\usepackage{amssymb}
\usepackage{float}
\usepackage{array}
\usepackage{mathtools}
\usepackage{pdfpages}
\newcommand{\myparagraph}[1]{\vspace{0.1em}\noindent\textbf{#1}}

\usepackage[breaklinks=true,bookmarks=false]{hyperref}

\iccvfinalcopy 

\def\iccvPaperID{****} 

\ificcvfinal\fi

\begin{document}

\title{ Neural Video Portrait Relighting in Real-time via Consistency Modeling}

\author{Longwen Zhang$^{1,2}$ \thanks{Equal contribution}
\hspace{0.5cm}
Qixuan Zhang$^{1,2}$  {\footnotemark[1]}
\hspace{0.5cm}
Minye Wu$^{1,3}$
\hspace{0.5cm}
Jingyi Yu$^1$
\hspace{0.5cm}
Lan Xu$^1$
\vspace{0.5cm}
\\
$^1$ShanghaiTech University 
\hspace{0.5cm}
$^2$Deemos Technology\\
$^3$University of Chinese Academy of Sciences\\
{\tt\small \{zhanglw2,zhangqx1,wumy,yujingyi,xulan1\}@shanghaitech.edu.cn}\\
{\tt\small \{zhanglw,zhangqx\}@deemos.com
}
}

\maketitle
\ificcvfinal\thispagestyle{empty}\fi
\thispagestyle{plain}
\pagestyle{plain}

\begin{abstract}
Video portraits relighting is critical in user-facing human photography, especially for immersive VR/AR experience.
Recent advances still fail to recover consistent relit result under dynamic illuminations from monocular RGB stream, suffering from the lack of video consistency supervision. 
In this paper, we propose a neural approach for real-time, high-quality and coherent video portrait relighting, which jointly models the semantic, temporal and lighting consistency using a new dynamic OLAT dataset.
We propose a hybrid structure and lighting disentanglement in an encoder-decoder architecture, which combines a multi-task and adversarial training strategy for semantic-aware consistency modeling.
We adopt a temporal modeling scheme via flow-based supervision to encode the conjugated temporal consistency in a cross manner.
We also propose a lighting sampling strategy to model the illumination consistency and mutation for natural portrait light manipulation in real-world.
Extensive experiments demonstrate the effectiveness of our approach for consistent video portrait light-editing and relighting, even using mobile computing.

\end{abstract}

\section{Introduction}
The past ten years have witnessed a rapid development of digital portrait photography with the rise of mobile cameras.
Relighting evolves as a cutting-edge technique in such portrait photography for immersive visual effects of VR/AR experience.
How to further enable consistent relit video results under challenging dynamic illumination conditions conveniently remains unsolved and has received substantive attention in both industry and academia.

\begin{figure}[tbp] 
	\centering 
	\includegraphics[width=0.8\linewidth]{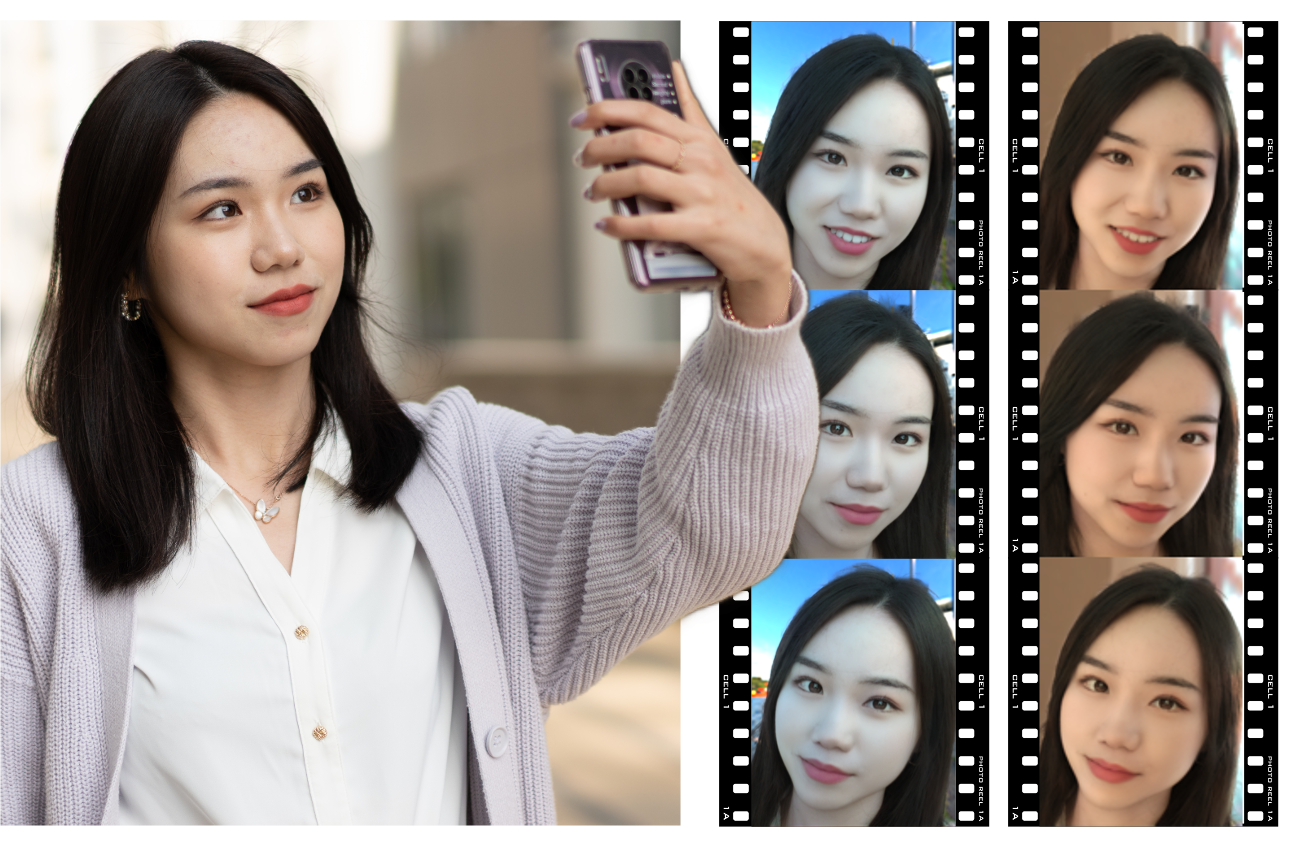} 
	\caption{Our approach achieves high-quality and consistent video portrait relighting under dynamic illuminations in real-time, using only mobile computing and monocular RGB video input.} 
	\label{fig:fig_1_teaser} 
	\vspace{-8pt} 
\end{figure}

For video portrait relighting, early solutions~\cite{debevec2000acquiring,wenger2005performance} rely on a sophisticated studio setup which is expensive and difficult to be deployed.
Modern approaches~\cite{shih2014style,song2017stylizing,Shu2017} further apply color or style transfer techniques to ease the hardware requirements. However, they still require two to four orders of magnitude more time than is available for interactive video application.
%
The recent learning techniques bring huge potential for human portrait modeling and relighting~\cite{Zollhoefer2018FaceSTAR,Zhou2019,Sun2019,nestmeyer2020learning,Wang2020} from only monocular RGB input.
In particular, the methods~\cite{shu2017neural,sengupta2018sfsnet} perform explicit neural inverse rendering but are limited to the low-quality face and Spherical Harmonics (SH) lighting models.
Recent methods~\cite{Zhou2019,Sun2019} remove the explicit inverse rendering by learning an efficient end-to-end mapping between the input headshots and relit ones, while the method~\cite{Wang2020} further models reflectance attributes explicitly to handle lighting effects like specular or shadow.
However, they still focus on single image input without modeling the temporal consistency for video portrait relighting, leading to severe jittery artifacts, especially under the challenging dynamic illuminations. 
Some recent mobile devices~\cite{Apple, Google} enables the ``Portrait Lighting'' mode for video editing of lighting conditions. Critically, they only modify existing illumination rather than relight the captured video into various scenes.

In this paper, we attack the above challenges and present a novel real-time and temporally coherent portrait relighting approach from only a monocular RGB video input, as illustrated in Fig.~\ref{fig:fig_1_teaser}.
Our approach jointly models the semantic, temporal and lighting consistency to enable realistic video portrait light-editing and relighting into new scenes with dynamic illuminations, whilst maintaining real-time performance even on portable device.

Generating such realistic and consistent video relit result in a real-time and data-driven manner is non-trivial.
From the data aspect, existing face datasets~\cite{cao2013facewarehouse,yang2020facescape,Wang2020} lack the video ground-truth supervision for consistency modeling.
Thus, we build up a high-quality dataset for video portrait relighting, consisting of 603,288 temporal OLAT (one light at a time) images at 25 frames per second (fps) of 36 actors and 2,810 environment lighting maps.
Our dynamic OLAT dataset is captured via a light stage setup with 114 LED light sources and a stationary 4K ultra-high-speed camera at 1000 fps.
From the algorithm side, we further propose a novel neural scheme for consistent video portrait relighting under dynamic illuminations. 
To maintain the real-time performance, we adopt the encoder-decoder architecture to each input portrait image similar to previous methods~\cite{Sun2019,Wang2020}.
Differently, we introduce a hybrid and explicit disentanglement for semantic-aware consistency, which self-supervises the portrait structure information and fully supervises the lighting information simultaneously in the bottleneck of the network.
Such disentanglement is further enhanced via multi-task training as well as an adversarial strategy so as to encode the semantic supervision and enable more realistic relighting.
Then, to utilize the rich temporal consistency in our dynamic OLAT dataset, a novel temporal modeling scheme is adopted between two adjacent input frames.
Our temporal scheme encodes the conjugated temporal consistency in a cross manner via flow-based supervision so as to model the dynamic relit effect.  
Finally, a lighting sampling based on Beta distribution is adopted, which augments the discrete environment lighting maps and generates a triplet of lighting conditions for adjacent input frames and the target output.
Our sampling scheme models the illumination consistency and mutation simultaneously for natural video portrait light-editing and relighting in the real-world.
To summarize, our main contributions include:
\begin{itemize} 
	\setlength\itemsep{0em}
	\item We present a real-time neural video portrait relighting approach, which faithfully models the video consistency for dynamic illuminations, achieving significant superiority to the existing state-of-the-art.
	
	\item We propose an explicit structure and lighting disentanglement, a temporal modeling as well as a lighting sampling schemes to enable realistic video portrait light-editing and relighting on-the-fly.
	
	\item We make available our dataset of 36 performers with 603,288 temporal OLAT images to stimulate further research of human portrait and lighting analysis. 
\end{itemize}

\section{Related Work} 
\myparagraph{Portrait Relighting.}
Debevec~\emph{et al.}~\cite{debevec2000acquiring} invent Light Stage to capture the reflectance field of human faces, which has enabled high-quality 3D face reconstruction and illuminations rendering, advancing the film's special effects industry. Some subsequent work has also achieved excellent results by introducing deep learning\cite{10.1145/3446328,meka2019deep,10.1145/3355089.3356571,meka2020deep, sun2020light}.
Obviously, this is not a product for individual consumers; thus various methods for single portrait relighting have been proposed. 
Several methods~\cite{xu2019deep,Bi_2020_CVPR,li2018learning,xu2018deep,tian2017depth} perform relighting on static objects.
Some work follows the pipeline of color transfer to achieve the relighting effects~\cite{chen2011face, shih2014style, song2017stylizing, Shu2017}, which usually needs another portrait image as the facial color distribution reference. 
Blanz~\emph{et al.}~\cite{blanz1999morphable} use a morphable model of faces that allows relighting by changing the directional lighting model parameters. 
With the advent of deep neural networks, recent approaches obtain various neural rendering tasks successfully~\cite{tewari2020state}.
Some methods~\cite{Zhou2019,liu2021relighting,sengupta2018sfsnet} adapt Spherical Harmonics (SH) lighting model to manipulate the illumination.
Several work~\cite{aldrian2012inverse, egger2018occlusion, wang2007face} jointly estimate the 3D face and SH~\cite{basri2003lambertian,ramamoorthi2001relationship} parameters and achieved relighting by recovering the facial geometry and modify the parameters of the SH lighting model.
Sevastopolsky~\emph{et al.}~\cite{sevastopolsky2020relightable} use point cloud to generate relightable 3D head portraits, while Tewari~\emph{et al.}~\cite{tewari2020pie} use GAN to manipulate the illumination.
Explicitly modeling the shadow and specular~\cite{Wang2020,nestmeyer2020learning} achieve excellent results in directional light source relighting. 
Mallikarjunr~\emph{et al.}~\cite{mbr_frf} take a single image portrait as input to predict OLAT(one-light-at-a-time) as Reflectance Fields, which can be relit to other lighting via image-based rendering.
Sun~\emph{et al.}~\cite{Sun2019} choose environment map as lighting model and use light stage captured OLAT data to generate realistic training data and train relighting networks in an end-to-end fashion. 
We also use the OLAT images to generate training data for portrait relighting. 
Differently, our approach enables real-time and consistent video portrait relighting under dynamic illuminations, with the aid of a new dynamic OLAT dataset.

\myparagraph{Temporal Consistency.}
Previous image relighting methods can be extended to videos if we directly treat every video frames as independent images. 
However, those methods will inevitably generate flicker results on relit videos. 
To suppress flicker results, several approaches have been developed for video style transfer tasks ~\cite{Litwinowicz1997,Lang2012,Bonneel2013,Ye2014,Intrinsic_Video,Bonneel2015,Ruder2016,Wang2018}. 
Specifically, Ruder~\emph{et al.}~\cite{Ruder2016} employed a temporal loss guided by optical flow for video style transfer, but the real-time computation of optical flows makes this approach slower. Vid2Vid~\cite{Wang2018} synthesised videos with temporal consistency by training a network to estimate optical flow and apply it on previously generated frames.
In this paper, we show that temporal consistency and portrait relighting can be simultaneously learned by a feed-forward CNN, which avoids computing optical flows in the inference stage.

\myparagraph{Video Relighting.}
Some methods \cite{wang2008video,5653028,5277971} use IR and LEDs around the screen and web camera to provide acceptable lighting for video conferencing.
Li~\emph{et al.}~\cite{li2014free} create free-viewpoint relighting video using multi-view reconstruction under general illumination, while Richardt~\emph{et al.}~\cite{richardt2012coherent} add video effects like relighting using RGBZ video cameras.
``The Relightables"~\cite{10.1145/3355089.3356571} proposes a hybrid geometric and machine learning reconstruction pipeline to generate high-quality relit video.
In contrast, our method does not require extra specific capturing equipment and enables real-time video portrait relighting using mobile computing.

\myparagraph{Face Dataset.}
Traditional face dataset usually takes 2D images in various lighting conditions~\cite{gross2010multi,lee2005acquiring,gao2007cas}. 
The controlled lighting conditions are easy to build but lack reflectance information for photo-realistic portrait relighting. 
With the development of face scanning and reconstruction techniques, 3D face datasets have been extended from only geometric~\cite{yin20063d,zhang2013high,savran2008bosphorus,cao2013facewarehouse,cheng20184dfab,cosker2011facs,zhang2014bp4d,yang2020facescape} to include reflectance channels~\cite{stratou2011effect,Wang2020}. 
However, the existing rendering scheme is difficult to avoid the Valley of Terror effect without manual modification. 
3D dataset still cannot achieve the realism in 2D face dataset or those using image-based rendering
Thus, various face OLAT datasets~\cite{mbr_frf,Sun2019,nestmeyer2020learning} have been proposed with light stage setup.
In contrast, we construct a new dynamic OLAT dataset through a a light stage setup and a 4K ultra-high-speed camera. 
Our high-quality dataset consists of 603,288 temporal OLAT imagesets at 25 fps of 36 subjects (18 females and 18 males) with various expressions and hairstyles.

\begin{figure}[t]

	\begin{center}
		\includegraphics[scale=0.28]{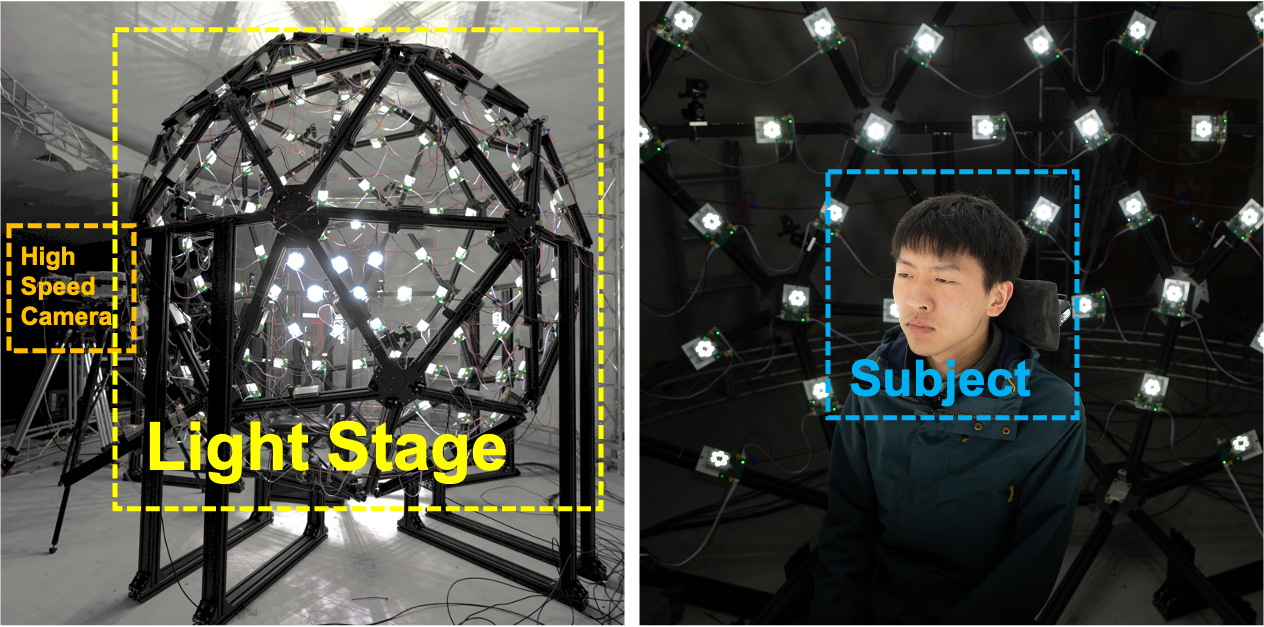}
	\end{center}
	\vspace{-10pt}
	\caption{Illustration of the capturing system for building our dynamic OLAT dataset.}
	\label{fig:Capture System}
	\vspace{-3mm}
\end{figure}

\section{Dynamic OLAT Dataset Overview}
Our goal is to naturally manipulate the lighting of a portrait RGB video captured in the wild into new environment lighting conditions while preserving the consistent structure and content.
To provide ground truth supervision for video consistency modeling, we build up a high-quality dynamic OLAT dataset.
As illustrated in Fig.~\ref{fig:Capture System}, our capture system consists of a light stage setup with 114 LED light sources and Phantom Flex4K-GS camera (global shutter, stationary 4K ultra-high-speed camera at 1000 fps), resulting in dynamic OLAT imageset recording at 25 fps using the overlapping method~\cite{wenger2005performance}.
Our dataset includes 603,288 temporal OLAT imagesets of 36 actors (18 females and 18 males) with 2810 HDR environment lighting maps~\cite{hold2019deep,gardner2017learning}, to provide high-quality and diverse video portrait relighting samples.
Besides, we apply a pre-defined light condition to each OLAT imageset to obtained a fully-illuminated portrait image.
Then, both the portrait parsing~\cite{zllrunning} and the optical flow~\cite{Nvidia} algorithms are applied to such a fully-illuminated stream to obtained ground truth semantics and correspondence supervisions.
Our dynamic OLAT dataset provides sufficient semantic, temporal and lighting consistency supervision to train our neural video portrait relighting scheme, which can generalize to in-the-wild scenarios.


\begin{figure}[]
	\begin{center}
		\includegraphics[width=0.7\linewidth]{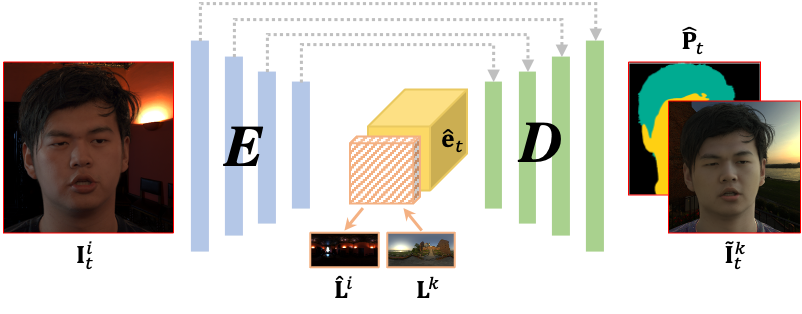}
	\end{center}
	\vspace{-13pt}
	\caption{Illustration of our video portrait relighting network based on encoder-decoder architecture for real-time inference.}
	\label{fig:fig4_method}
		
\end{figure}

\begin{figure*}
\begin{center}
\includegraphics[width=0.9\linewidth]{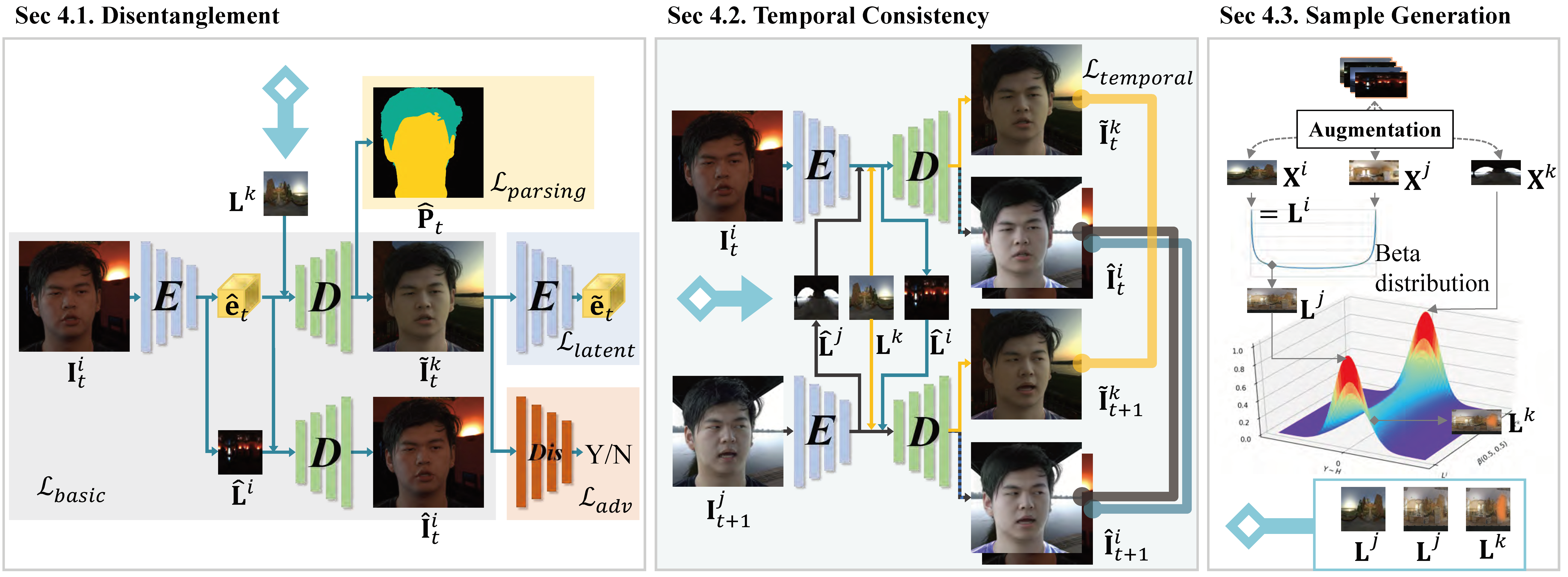}
\end{center}
\vspace{-12pt}
\caption{The training pipeline of our approach. It consists of a structure and lighting disentanglement (Sec.~\ref{sec:4.1}), a temporal consistency modeling (Sec.~\ref{sec:4.2}) and a lighting sampling  (Sec.~\ref{sec:4.3}), so as to generate consistent video relit results from a RGB stream in real-time.}
\label{fig:pipeline}
\end{figure*}

\section{Neural Video Portrait Relighting}
Our scheme obtains real-time and coherent video relit results from only mobile computing and monocular RGB steam, using our dynamic OLAT dataset, as shown in Fig.~\ref{fig:pipeline}. 
First, a hybrid disentanglement for portrait structured and lighting is introduced for semantic-aware consistency modeling (Sec.~\ref{sec:4.1}).
Then, a temporal modeling is adopted to utilize the flow-based supervision for dynamic relit effects (Sec.~\ref{sec:4.2}).
We also introduce a lighting sampling strategy to model the illumination consistency and mutation for natural portrait lighting manipulation in real-world scenarios (Sec.~\ref{sec:4.3}).

\myparagraph{Notations}.
To achieve the real-time performance, we adopt the encoder-decoder architecture to the portrait stream sequentially similar to previous methods~\cite{Sun2019,Wang2020}, as illustrated in Fig.~\ref{fig:fig4_method}.
Given an input image $\mathbf{I}^i_t$ at $t$-th frame and a desired target lighting $\mathbf{L}^k$ (environment lighting map), our network $\Phi$ predicts both a target portrait image $\mathbf{\tilde{I}}^k_t$ lit by the $\mathbf{L}^k$ and a corresponding semantic mask $\hat{\mathbf{P}_t}$: 
\begin{equation}
  \mathbf{\tilde{I}}^k_t, \mathbf{\hat{L}}^i, \mathbf{\hat{P}}_t = \Phi(\mathbf{I}^i_t, \mathbf{L}^k), 
  \label{eqn:network}
\end{equation}
where $\mathbf{\hat L}^i$ is the regressed lighting of the input image $\mathbf{I}^i_t$.
Specifically, the encoder $\Phi_{enc}$ encodes the input image into both the lighting $\mathbf{L}^i$ and a portrait structure latent code $\mathbf{\hat e}_t$:
\begin{equation}
\begin{split}
    \mathbf{\hat{L}}^i, \mathbf{F}^i_t, \mathbf{\hat e}_t &= \Phi_{enc}(\mathbf{I}^i_t),
\end{split}
\end{equation}
where $\mathbf{F}^i_t$ is the output of skip connections.
%
%
Similarly, the corresponding decoder $\Phi_{dec}$ is formulated as:
\begin{equation} \label{equ:decoder}
\begin{split}
    \mathbf{\hat{I}}^i_t, \mathbf{\hat{P}}_t &= \Phi_{dec}(\mathbf{\hat L}^i, \mathbf{F}^i_t, \mathbf{\hat e}_t), 
\end{split}
\end{equation}
where we use predicted illumination $\mathbf{\hat L}^i$ to relit itself. 
By replacing the lighting with a known one $\mathbf{L}^k$ in the decoder, we can obtain a relit portrait image $\mathbf{\tilde I}_t^k$ corresponding to  $\mathbf{L}^k$. 
Note that we use tilde and hat notations for the images relit by known illumination or the predicted one, respectively.

\subsection{Structure and Lighting Disentanglement} \label{sec:4.1}
The core of realistic relighting is the reliable disentanglement of the portrait structure information $\mathbf{e}$ and the lighting condition $\mathbf{L}$ in the bottleneck of our network in Eqn.~\ref{eqn:network}.
To this end, we adopt a hybrid disentanglement scheme combing with multi-task training and adversarial training strategy to model the semantic consistency for realistic relighting.

Similar to the method~\cite{Sun2019}, during training we optimize the following basic loss for disentanglement, which minimizes the photometric error and illumination distance between the prediction and the ground truth from our dataset:
\begin{equation}
\begin{split}
\mathcal{L}_{basic} &=\frac{1}{2}\|\log(1 + {\mathbf{L}}^i)-\log( 1+ {\mathbf{\hat L}}^i)\|_2^2 \\
+&\|\mathbf{M}_t\odot({\mathbf{I}}_t^k-{\mathbf{\tilde I}}_t^{k})\|_1 + \|\mathbf{M}_t\odot({\mathbf{I}}_t^i-{\mathbf{\hat I}}_t^i)\|_1, 
\end{split}
\end{equation}
where $\mathbf{M}_t$ is a portrait foreground mask of the $t$ corresponding frame from face parsing and $\odot$ is element-wise multiplication.
However, only using this basic scheme fails to provide supervision on the structure latent code $\mathbf{e}$ and encode the rich semantic consistency in our dynamic OLAT dataset. 
Thus, we introduce the following strategies for more reliable disentanglement and more consistent relighting. 

\myparagraph{Structure Self-supervision.}
Recall that the method~~\cite{Sun2019} treats the latent space as lighting map and only relies on the feature maps $\mathbf{F}^i_t$ for modeling the portrait structure information.
Differently, we utilize a separated structure-wise latent code $\mathbf{\hat e}_t$, which has larger receptive field in the encoder to represent the global context information of the portrait image.
Thus, we design a novel self-supervised scheme for such structure-wise latent code by applying the encoder recurrently to the relit output and enforces the consistency between the structure codes for further disentanglement. We formulate it as follows:
\begin{equation}
\begin{split}
\mathcal{L}_{latent} & =\|\mathbf{\hat{e}}_t-{\mathbf{\tilde e}}_t\|_2^2,
\end{split}
\label{eq:latent}
\end{equation}
where $\mathbf{\tilde{e}}_t$ is from the recurrent output of $\Phi_{enc}(\mathbf{\tilde{I}}^k_t)$. Here we utilize encoder $\Phi_{enc}$ to encode the relit image $\mathbf{\tilde{I}}^{k}_t$ with the target light $\mathbf{L}^k$ and verify its global structure latent code with the original one to enhance structure consistency.

\myparagraph{Semantics-aware Multi-Task Learning.}
Human face skin with scattering effects owns different reflectance distribution from other materials like hair. 
Treating all the portrait pixels uniformly will cause strong artifacts in relighted results without maintaining the semantic consistency. 
To this end, we design the multi-task decoder $\Phi_{dec}$ which aims to restore both the relighted image $\mathbf{\hat{I}}^o_t$ and the semantic portrait mask $\mathbf{P}_t$ under the supervision from our dataset. 
Such parsing loss with binary cross-entropy metrics is formulated as:
\begin{equation}
\begin{split}
\mathcal{L}_{parsing} & =-\Big({{\mathbf{P}}}_t\odot\log {\mathbf{\hat P}}_t+(1-{ {\mathbf{P}}}_t)\odot\log (1-{\mathbf{\hat  P}}_t)\Big).
\end{split}
\end{equation}
By predicting semantic portrait mask $\mathbf{P}_t$, we enforce both encoder and decoder networks to be aware of the semantic information of human portrait image. 
Hence the networks can implicitly models the semantic consistency for more realistic relighting of various portrait regions.

\myparagraph{Adversarial Training.} 
To further reinforcing portrait image details, we also introduce a discriminator network $\Phi_{D}$, which has the same architecture as DCGAN's~\cite{radford2015unsupervised}. 
We adopt Wasserstein GAN~\cite{arjovsky2017wasserstein} strategies to the proposed discriminator for a stable training process. 
Specifically, we remove the activation function of $\Phi_{D}$'s final layer and apply the weight clipping method during training. 
Then, the adversarial losses is formulated as:
\begin{equation}
\begin{split}
\mathcal{L}_{adv_D} &= -\Phi_{D}({\mathbf{I}}_t^i,{\mathbf{I}}_t^k,{\mathbf{L}}^k)+\Phi_{D}({\mathbf{I}}_t^i,{\mathbf{\tilde I}}_t^k,{\mathbf{L}}^k)\\
\mathcal{L}_{adv_G} &= -\Phi_{D}({\mathbf{I}}_t^i,{\mathbf{\tilde I}}_t^k,{\mathbf{L}}^k),
\end{split}
\end{equation}
where $\mathcal{L}_{adv_D}$ is only for updating the discriminator, and $\mathcal{L}_{adv_G}$ is only for updating the decoder; $\mathcal{L}_{adv} = \mathcal{L}_{adv_D} + \mathcal{L}_{adv_G}$.
Here, the discriminator takes a triplet as input, including a source image, a relit image, and the corresponding light condition ${\mathbf{L}}^k$, which estimates the Wasserstein distance between the real image distribution and the relit image distribution. 
Note that the structure of the source image and ${\mathbf{L}}^k$ are essential cues for such distance measurement.

\subsection{Temporal Consistency Enhancement}  \label{sec:4.2}
Previous single-image relighting approaches lack the explicit temporal modeling in terms of portrait motions or dynamic illuminations, leading to flickering artifacts for video applications. 
Thus, we propose a rescue by utilizing the rich temporal consistency supervision in our dynamic OLAT dataset.
Specifically, for two adjacent training samples at timestamps $t$ and $t+1$, we obtain the forward flow $f_{t,t+1}(\cdot)$ and the backward flow $f_{t+1,t}(\cdot)$ from our continues OLAT imagesets, where $f_{a,b}(\cdot)$ warps images from time $a$ to time $b$. 
Note that our high-quality OLAT imagesets at 25 fps ensures the accuracy of such supervision based on optical flow for self-supervised verification in a cross manner.
	
For balancing the lighting distribution between two frames, we also introduce two conjugated illuminations in our training scheme. 
Given two adjacent frames $\mathbf{I}_t^i$ and $\mathbf{I}_{t+1}^j$, we relit both of them using our network according to the predicted lighting conditions $\mathbf{\hat L}^i$, $\mathbf{\hat L}^j$, as well as the target illumination $\mathbf{L}^k$  to obtain corresponding relit images at both frames. 
Thus, our temporal loss is formulated as:
	\begin{equation}
	\begin{split}
	&\mathcal{L}_{temporal} = \|f_{t,t+1}({\mathbf{\tilde I}}_t^k)-\mathbf{\tilde I}_{t+1}^k\|_1 +\|f_{t+1,t}({\mathbf{\tilde I}}_{t+1}^k)-{\mathbf{\tilde I}}_{t}^k\|_1 \\
	&+\sum_{z\in\{i,j\}}\big(\|f_{t,t+1}({\mathbf{\hat I}}_t^z)-{\mathbf{\hat I}}_{t+1}^z\|_1 + \|f_{t+1,t}({\mathbf{\hat I}}_{t+1}^z)-{\mathbf{\hat I}}_{t}^z\|_1\big),\\
	\end{split}
	\end{equation}
which encodes the conjugated temporal consistency in a cross manner via flow-based supervision so as to model the dynamic relit effect.

\subsection{ Lighting Conditions Sampling}  \label{sec:4.3}
Note that the discrete environment lighting maps in our dynamic OLAT dataset still cannot models the illumination consistency and mutation for real-time video relighting scenarios.
Thus, we introduce a novel lighting sampling scheme during training to generate a triplet of lighting conditions for adjacent input frames and the target output, which enhances the illumination consistency for natural portrait relighting and lighting manipulation.
 
Similar to previous work~\cite{Sun2019,Wang2020}, the lighting condition $\mathbf{L}$ is represented as a flattened latitude-longitude format of size $16\times16$ with three color channels from the environment lighting map and we adopt the same environment map re-rendering with random rotation to augment the lighting conditions in our dataset, which forms a uniform lighting condition sampling distribution $\mathcal{G}$.
To model illumination mutation, we further design a lighting condition distribution $\mathcal{H}$, where we randomly sample one to three point light sources with uniformly random colors outside a unit sphere.
The maximum distance from the light source to the sphere is limited by $1.5$ in order to produce reasonable illumination. 
Then, we project these light sources on a sphere according to the Lambertian reflectance model~\cite{basri2003lambertian} to form an environment map and corresponding lighting condition.

For each training sample, we generate three illuminations using different sampling strategies, including $\mathbf{L}^i$ and $\mathbf{L}^j$ to generate the adjacent images $\mathbf{I}^{i}_{t}$ and  $\mathbf{I}^{j}_{t+1}$, as well as the $\mathbf{L}^k$ for the target image $\mathbf{I}^{k}$, which is formulated as:
	\begin{equation}
	\begin{aligned}
	\mathbf{L}^i &= \mathbf{X}^i \\
	\mathbf{L}^j &= \beta_1\mathbf{L}^i + (1-\beta_1)\mathbf{X}^j \\
	\mathbf{L}^k &= \beta_2\mathbf{L}^j + (1-\beta_2)\mathbf{X}^k + \mathbf{Y},
	\end{aligned}
	\end{equation}
	where $\mathbf{X}^i, \mathbf{X}^j, \mathbf{X}^k \sim \mathcal{G}$, $\mathbf{Y} \sim \mathcal{H}$ and $\beta_1, \beta_2$ are sampled from a Beta distribution $Beta(0.5,0.5)$. 
Here, the Beta distribution drastically diversifies the lighting combination for modeling the illumination consistency and mutation simultaneously and enhancing the generalization ability of our network.
Conceptually, $\mathbf{L}^i$ and $\mathbf{L}^j$ have similarity, contributing to the coverage of temporal loss training. And $\mathbf{L}^k$ provides challenging illumination examples to enhance the lighting manipulation ability of our approach. 


\subsection{Implementation Details}
We utilize our dynamic OLAT dataset to train our video relighting network.
Our training dataset consists of 536,256 temporal OLAT imagesets of 32 actors lasting for 188.16 seconds.
The remaining OLAT imagesets of the other four actors unseen during training are taken as the test dataset. 
Note that we also augment the dataset using random cropping and resizing to add more challenging incomplete portraits to enhance the generalization ability of our network. 
During training our total loss is formulated as follows:
\begin{equation}
	\begin{split}
	\mathcal{L}  = \lambda_1\mathcal{L}_{basic} +& \lambda_2\mathcal{L}_{latent} + \lambda_3\mathcal{L}_{parsing}  \\
	&+ \lambda_4\mathcal{L}_{temporal} + \lambda_5\mathcal{L}_{adv},
	\end{split}
\end{equation}
where the weights for each term are set to be $1.0$ in our experiments. 
Since we utilize Wasserstein GAN in our approach, our network's parameters are optimized by RMSprop algorithm with a learning rate of $5e^{-5}$. 
Besides, we also clamp parameters of discriminator into a range of $[-0.01,0.01]$ and adopt the progressive training strategy.

\begin{figure*}[!h]
	\begin{center}
		\includegraphics[width=16.5cm]{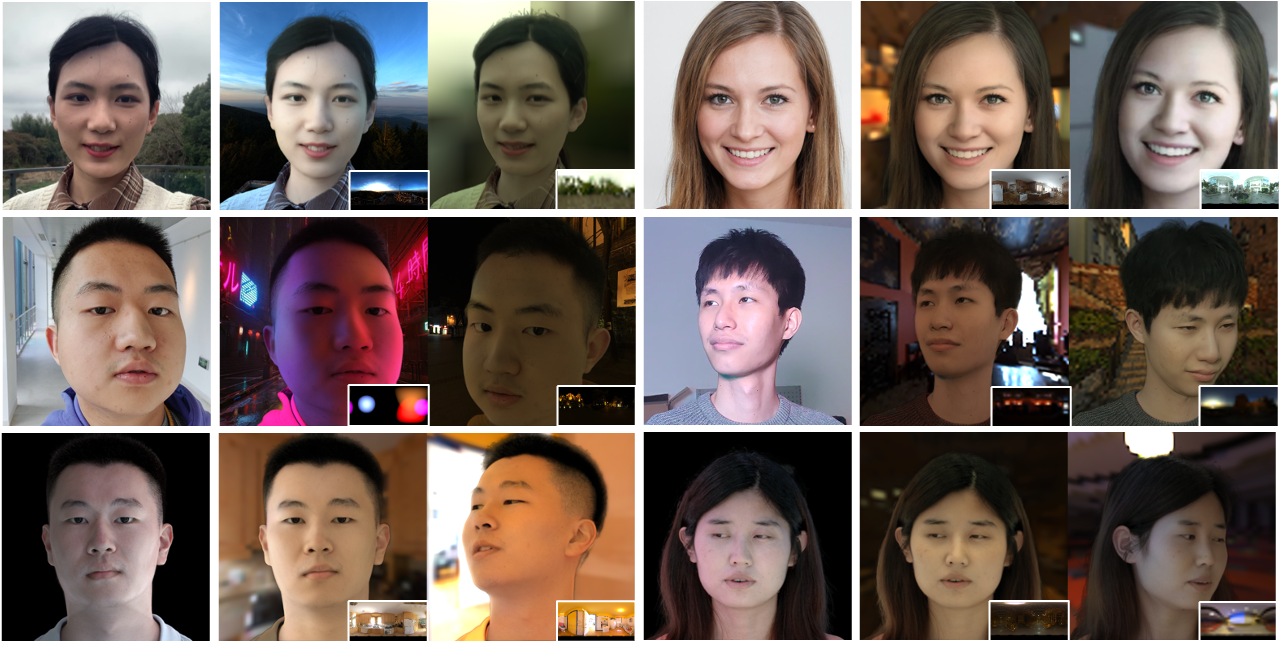}
	\end{center}
	\vspace{-10pt}	
	\caption{Our relighting results under dynamic illuminations. Each triplet includes the input frame and two relit result examples.}	
	\label{fig:gallery}
\end{figure*}

\begin{figure*}
	\setlength{\tabcolsep}{4pt}
	\newcommand{\figsize}{2.15cm}
	\newcommand{\ig}[1]{\includegraphics[width=\figsize,height=\figsize]{./fig/all/#1.png}}
	\begin{center}

	\begin{tabular}{
	>{\centering}p{\figsize}
	>{\centering}p{\figsize}
	>{\centering}p{\figsize}
	>{\centering}p{\figsize}
	>{\centering}p{\figsize}
	>{\centering}p{\figsize}
	>{\centering}p{\figsize}
	}
	 \multicolumn{7}{c}{\includegraphics[width=17cm]{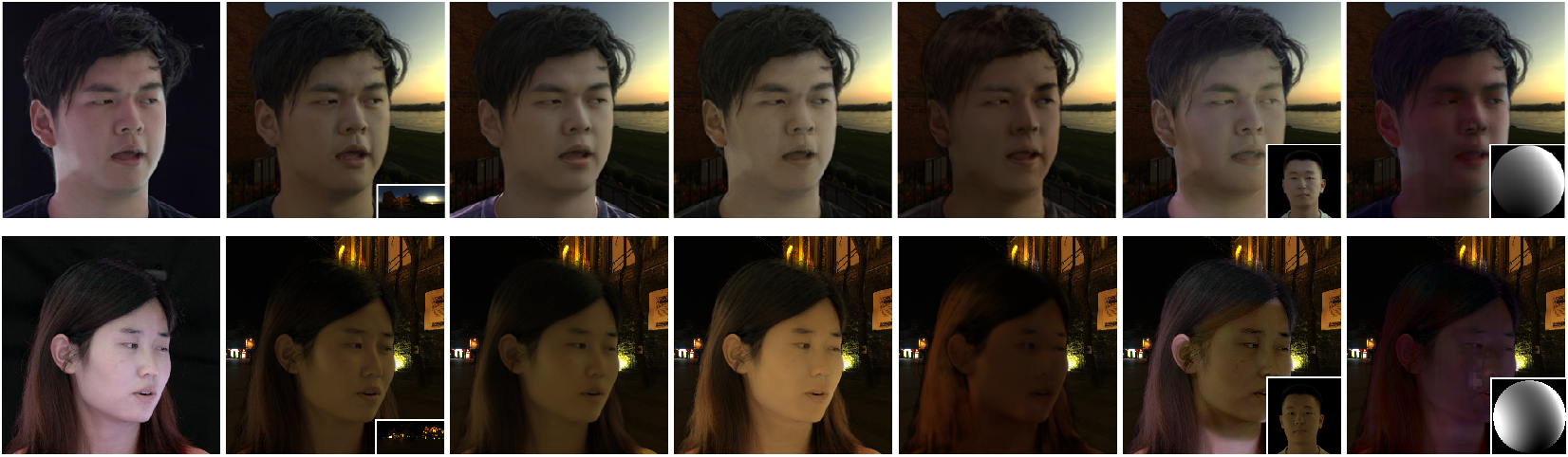}}
	 \\
	 Input & Ground Truth & Ours & SIPR & EMRCM & MTP & DPR
	\end{tabular}
	\end{center}
	\vspace{-5pt}
	\caption{Qualitative comparisons of relit results on our dynamic OLAT dataset. Our approach achieves more realistic relighting.}
	\label{fig:comparison}
\end{figure*}

\newpage

\begin{figure*}
	\setlength{\tabcolsep}{4pt}
	\newcommand{\figsize}{2.15cm}
	\newcommand{\ig}[1]{\includegraphics[width=\figsize,height=\figsize]{./fig/all/#1.png}}
	\begin{center}
		\begin{tabular}{
				>{\centering}p{\figsize}
				>{\centering}p{\figsize}
				>{\centering}p{\figsize}
				>{\centering}p{\figsize}
				>{\centering}p{\figsize}
				>{\centering}p{\figsize}
				>{\centering}p{\figsize}
			}
			\multicolumn{7}{c}{\includegraphics[width=14.5cm]{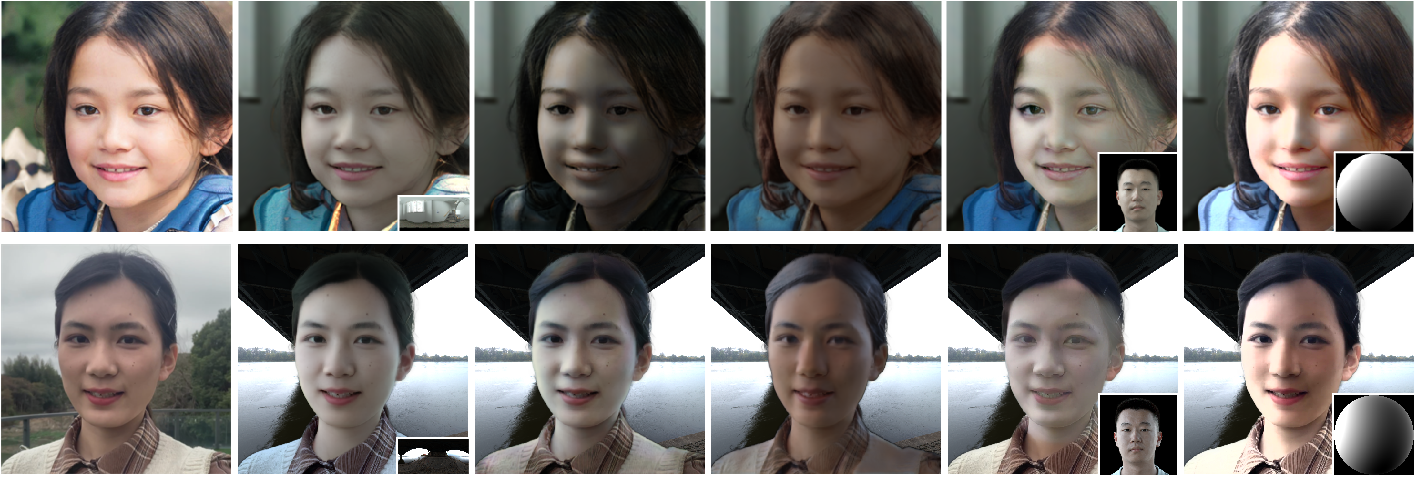}}
			\\
			Input & Ours & SIPR & EMRCM & MTP & DPR
		\end{tabular}
	\end{center}
	\vspace{-10pt}
	\caption{Qualitative comparisons of relit results on online or in-the-wild sequences. We obtain more natural results on unseen performers.}
	\label{fig:comparison_wild}
\end{figure*}

\section{Experiments}
Here we evaluate our method in various challenging scenarios. 
We run our experiments on a PC with Intel i7-8700K CPU, $32$GB RAM, and Nvidia RTX 3080 GPU, where our approach generates high-quality 512$\times$512 relit results at 111 fps (15 fps on the iPhone 12 device). 
Fig.~\ref{fig:gallery} demonstrates several results of our approach, which can generate consistent video relit results of both in-the-wild sequences and the one from our dataset with challenging illuminations.

\subsection{Comparison}
We compare our approach against existing state-of-the-art methods, including Single Image Portrait Relighting (\textbf{SIPR})~\cite{Sun2019}, the one via Explicit Multiple Reflectance Channel Modeling (\textbf{EMRCM})~\cite{Wang2020}, the one based on Mass Transport Approach (\textbf{MTP})~\cite{Shu2017} and Deep Portrait Relighting (\textbf{DPR})~\cite{Zhou2019}.
Note that we re-implement SIPR~\cite{Sun2019} and train it using our dataset for a fair comparison.
Fig.~\ref{fig:comparison} and Fig.~\ref{fig:comparison_wild} provide the qualitative comparison on both our dynamic OLAT dataset and online or in-the-wild sequences, respectively.
Note that our approach achieves significantly more realistic relit results under challenging illuminations by modeling the video consistency.

\begin{figure}[t]
	\centering
	\includegraphics[width=\columnwidth]{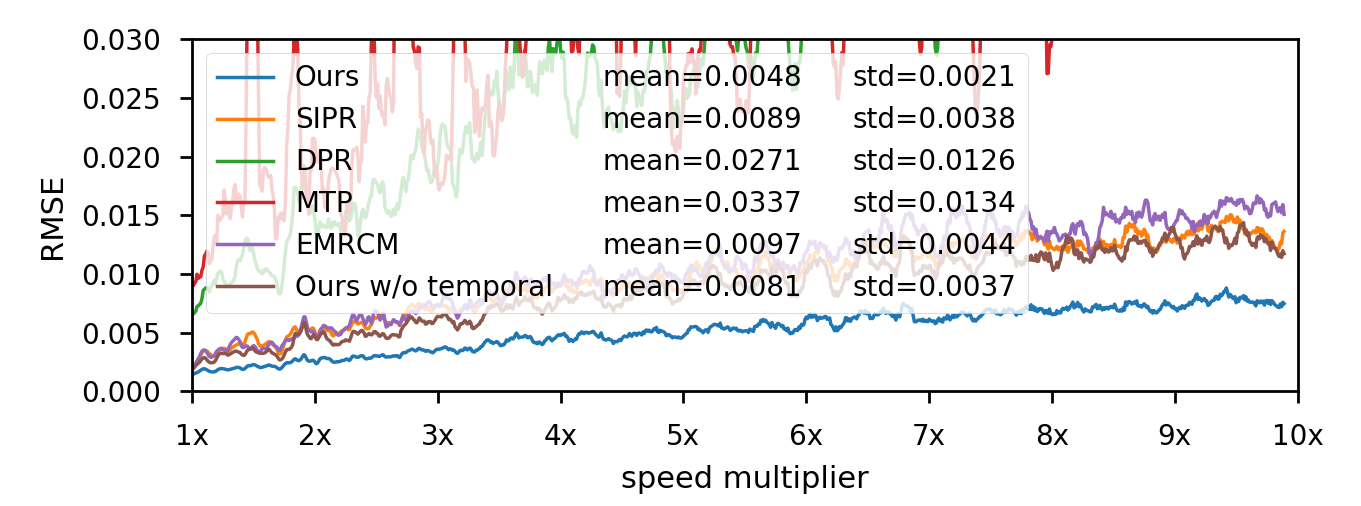}
	\vspace{-10pt}
	\caption{Quantitative comparison in terms of handling fast changing illuminations. Our approach consistently outperforms other baselines under various lighting speed up factors. 
	}
	\label{fig:10x}
\end{figure}

Then, we utilize our testing set with ground truth for quantitative comparison.
Similar to previous methods~\cite{Sun2019, Wang2020}, we adopt the \textbf{RMSE}, \textbf{PSNR}, and \textbf{SSIM} as metrics. 
Note that the output values are normalized to $[0,1]$, and only the valid portrait regions are considered.
As shown in Tab.~\ref{table:comparison}, our approach consistently outperforms the baselines in terms of these metrics above, illustrating the effectiveness of our approach for consistent video portrait relighting.
We further compare against baselines under dynamic illuminations. 
Thus, we synthesize 1000 frames with the static performer and changing lighting conditions using various speed-up factors from 1 to 10.  
Then, we relit the sequence into static lighting and calculated the average RMSE of adjacent output frames as the error metric for the jittery artifacts.
As shown in Fig.~\ref{fig:10x}, the error of our approach glows much slower consistently compared to others, which illustrates our superiority to handle dynamic illuminations.

\begin{table}[t]
	\footnotesize
	\newcommand{\tabincell}[2]{\begin{tabular}{@{}#1@{}}#2\end{tabular}}
	\renewcommand{\arraystretch}{1.2}	
	\centering
	\begin{tabular}{cccccc}
		\hline\hline
		Method & RMSE & PSNR & SSIM
		\\
		\hline\hline
		{SIPR}  &0.0974 &20.6542 &0.8901\\\hline
		{EMRCM} &0.0766 &22.7197 &0.8748\\\hline
		{MTP}   &0.0902 &21.9535 &0.8775\\\hline
		{DPR}   &0.1080 &20.8042 &0.8593\\\hline
		\textbf{Ours}  &\textbf{0.0349} &\textbf{30.6110} &\textbf{0.9584}\\\hline
	\end{tabular}
	\vspace{3pt}
	\caption{Quantitative comparison on our dynamic OLAT dataset.}
	\label{table:comparison} 
\end{table}


\subsection{Evaluation} 

\begin{figure}[t]
	\centering
	\setlength{\tabcolsep}{0.5pt}
	\newcommand{\figsize}{1.9cm}
	\newcommand{\ig}[1]{\includegraphics[width=\figsize,height=\figsize]{./fig/evaluation/GAN/#1.png}}
	\begin{tabular}{cccccccc}
		& \ig{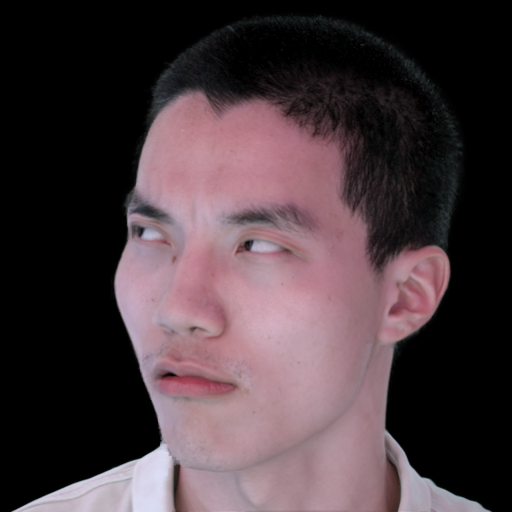} & \ig{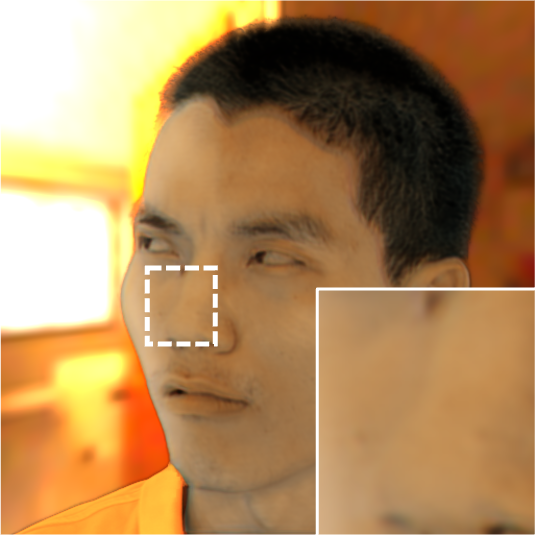} & \ig{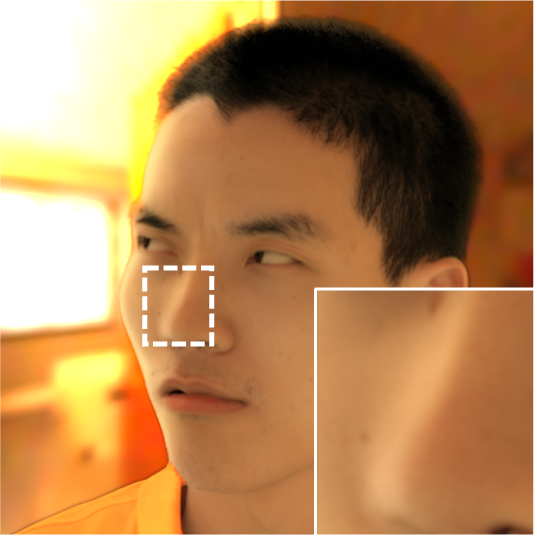}
		\\
		& \ig{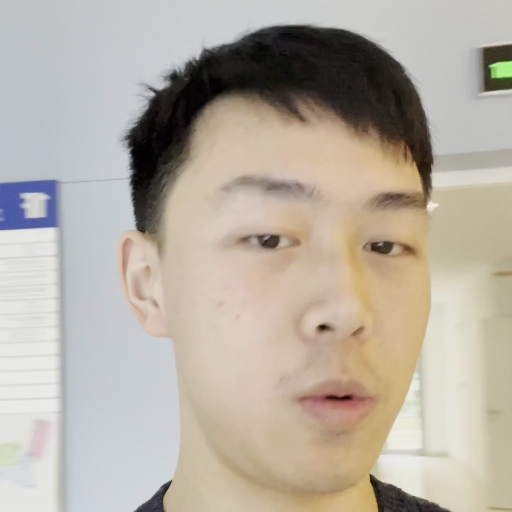} & \ig{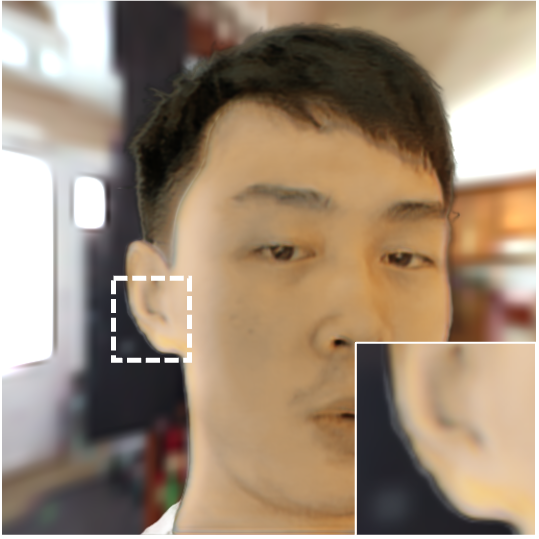} & \ig{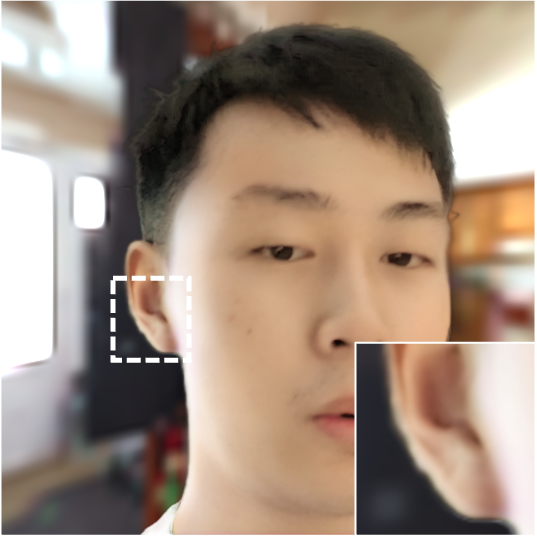}
		\\
		& Input & w/o structure & Ours
	\end{tabular}
	\caption{Qualitative evaluation of structure self-supervison. Our full pipeline achieve sharper relighting with fine structured details.}
	\label{fig:without_content}
\end{figure}

\begin{figure}[t]
	\centering
	\setlength{\tabcolsep}{0.5pt}
	\newcommand{\figsize}{1.9cm}
	\newcommand{\ig}[1]{\includegraphics[width=\figsize,height=\figsize]{./fig/evaluation/parsing/#1.png}}
	\begin{tabular}{cccccccc}
		& \ig{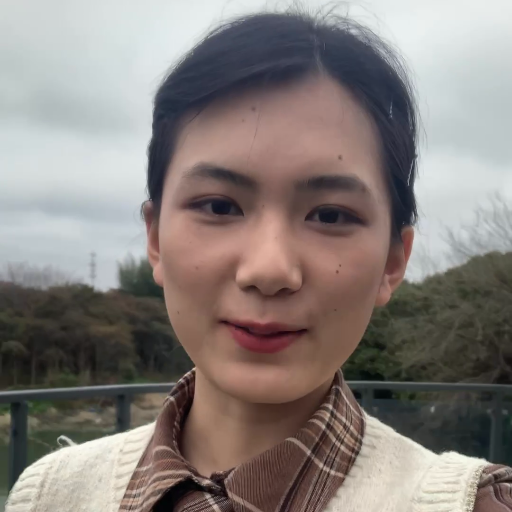} & \ig{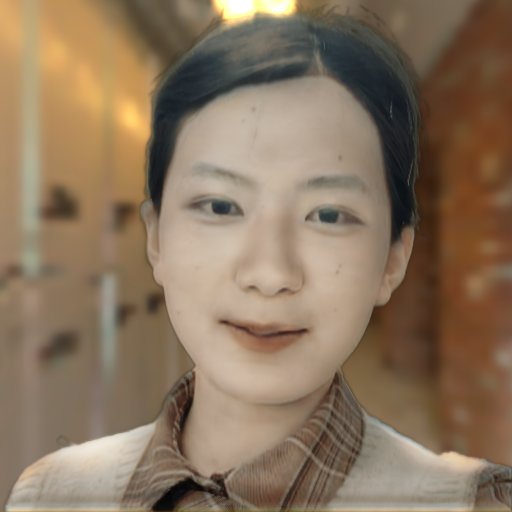} & \ig{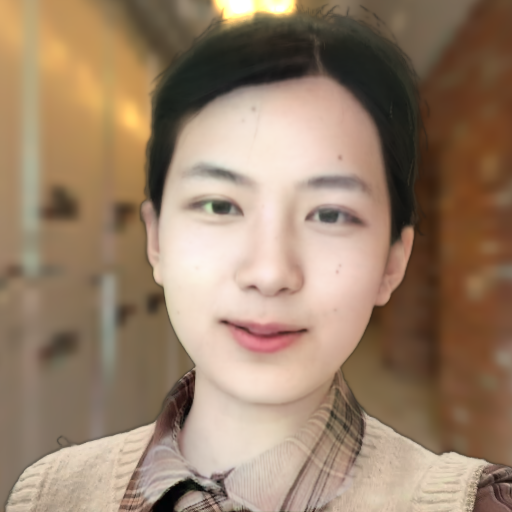}
		\\
		& \ig{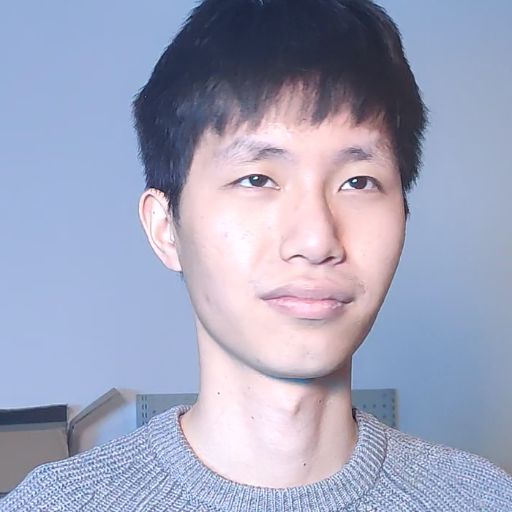} & \ig{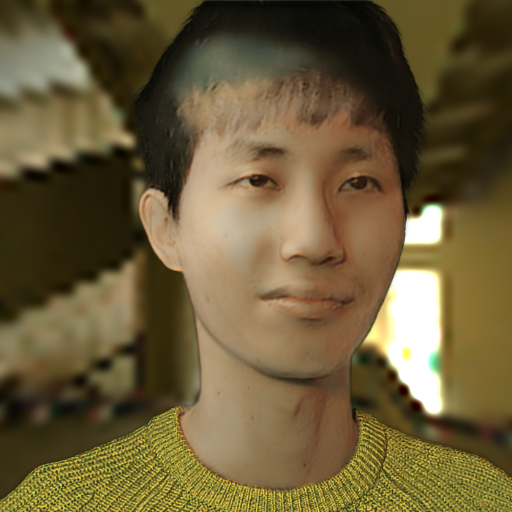} & \ig{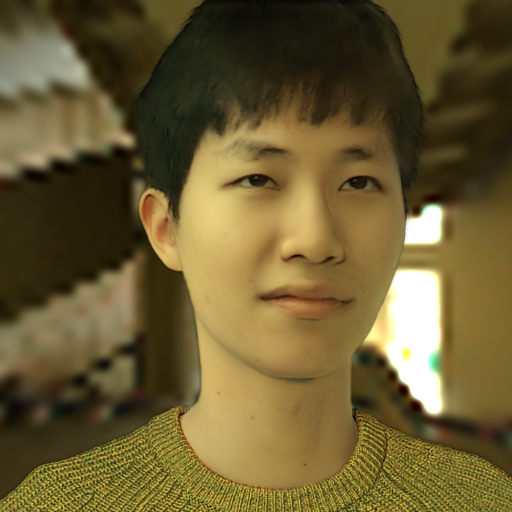}
		\\
		& Input & w/o enhance & Ours
	\end{tabular}
	\caption{Qualitative evaluation of disentangle enhancement. Our scheme models the semantic consistency for realistic relighting.}
	\label{fig:without_parsing}
	\vspace{-10pt}
\end{figure}

\myparagraph{Hybrid disentanglement.}
Here we evaluate our hybrid scheme for structure and lighting disentanglement.
Let \textbf{w/o~structure} denote the variation of our approach without the self-supervision of portrait structure in Eqn.~\ref{eq:latent}, and \textbf{w/o~enhance} denote the variation without the disentangle enhancement of multi-task and adversarial strategy.
As shown in Fig.~\ref{fig:without_content}, our scheme with structure self-supervision enables more accurate disentanglement for sharper realistic results.
The qualitative evaluation in Fig.~\ref{fig:without_parsing} further illustrates that our multi-task and adversarial training strategy encodes the semantic consistency for more realistic relighting.

\begin{figure}[t]
	\centering
	\setlength{\tabcolsep}{0.5pt}
	\newcommand{\figsize}{1.9cm}
	\newcommand{\ig}[1]{\includegraphics[width=\figsize,height=\figsize]{./fig/evaluation/timing/#1.png}}
	\begin{tabular}{cccccccc}
		& \ig{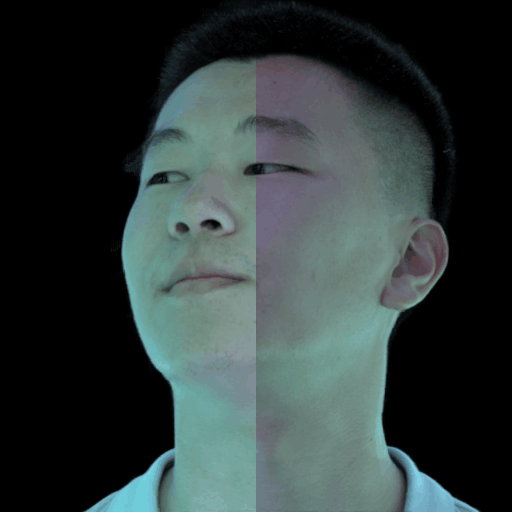}
		& \ig{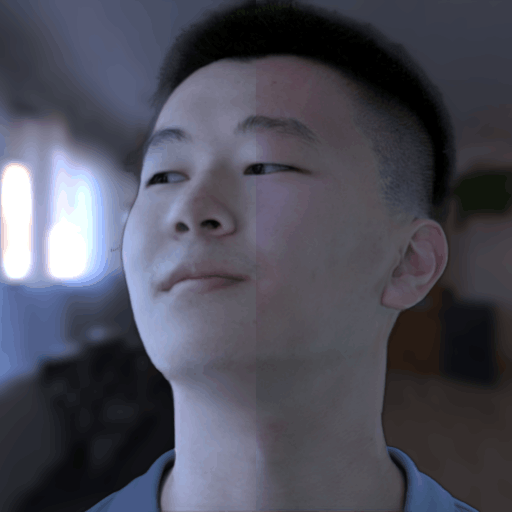} & \ig{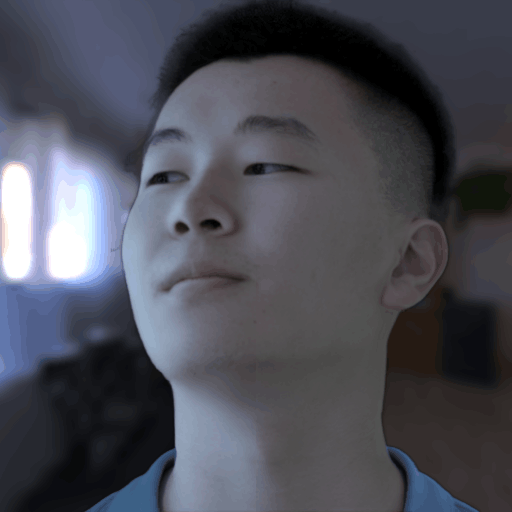}
		\\
		& Input 
		& w/o temporal & Ours
	\end{tabular}
	\includegraphics[width=\columnwidth]{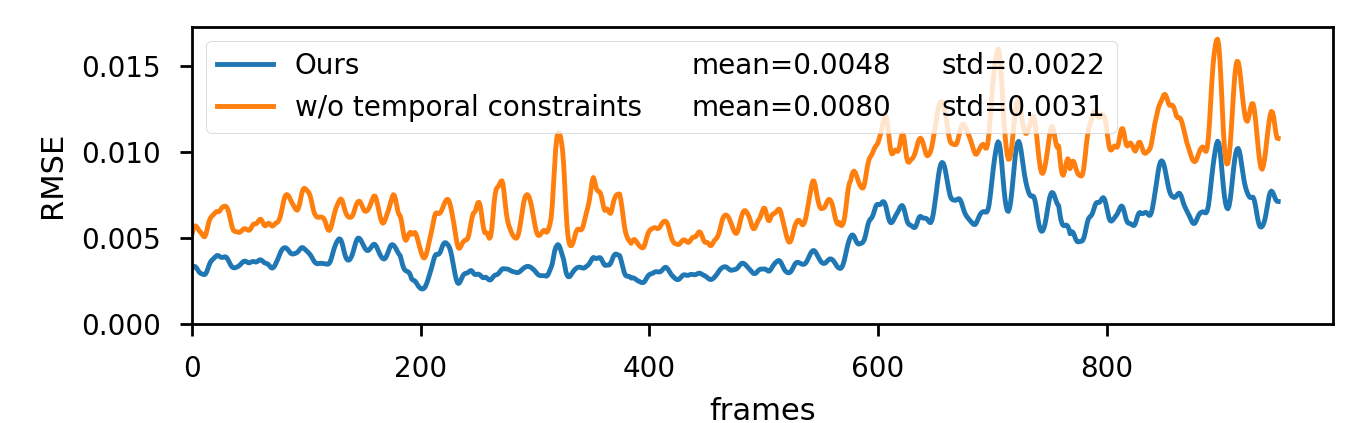}
	\caption{Evaluation of temporal modeling. Top: two relit examples of a static portrait from changing illuminations into a target lighting condition. Down: the corresponding error curve.}
	\label{fig:without_temporal}
\end{figure}

\myparagraph{Temporal modeling.}
Here we compare against our variation without temporal consistency modeling, denoted as \textbf{w/o~temporal}.
Similar to the comparison under dynamic illuminations, we relit the same synthesized sequence with static performer and changing illuminations into a target lighting condition for thorough evaluation. 
As shown in Fig.~\ref{fig:without_temporal}, our approach with temporal modeling achieves more temporal consistent results both qualitatively and quantitatively. 
We also provide quantitative evaluation under various lighting speed-up factors in Fig.~\ref{fig:10x}.
These results illustrate the effectiveness of our scheme to utilize temporal consistency.

\begin{table}[t]
	\footnotesize
	\newcommand{\tabincell}[2]{\begin{tabular}{@{}#1@{}}#2\end{tabular}}
	\renewcommand{\arraystretch}{1.2}	
	\centering
	\begin{tabular}{cccccc}
		\hline\hline
		Method & RMSE & PSNR & SSIM
		\\
		\hline\hline
		w/o content & 0.0549 & 25.5496 & 0.9021
		\\
		w/o temporal & 0.0404 & 28.6403 & 0.9510
		\\
		w/o parsing & 0.0680 & 23.6189 & 0.9170
		\\
		w/o sampling & 0.0616 & 24.5142 & 0.9223
		\\
		\textbf{Ours} & \textbf{0.0349} & \textbf{30.6110} & \textbf{0.9584}
		\\
		\hline
	\end{tabular}
	\vspace{3pt}
	\caption{Quantitative evaluation on synthesis sequences.}
	\label{table:evaluation} 
	\vspace{-10pt}
\end{table}

\begin{figure}[t]
	\centering
	
	\setlength{\tabcolsep}{0.5pt}
	\newcommand{\figsize}{1.9cm}
	\newcommand{\ig}[1]{\includegraphics[width=\figsize,height=\figsize]{./fig/evaluation/sampling/#1.png}}
	
	\begin{tabular}{cccccccc}
		\\
		\rotatebox{90}{w/o sampling} & \ig{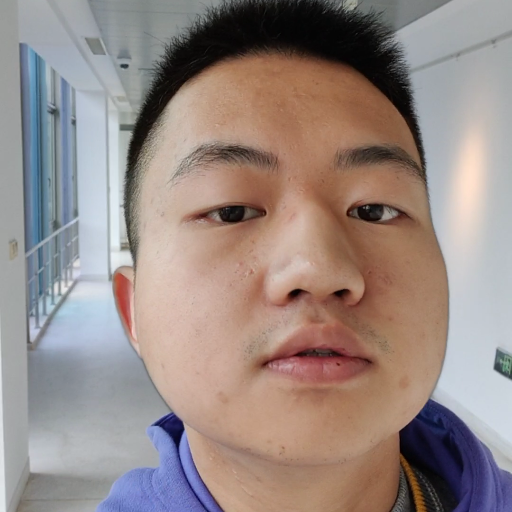} & \ig{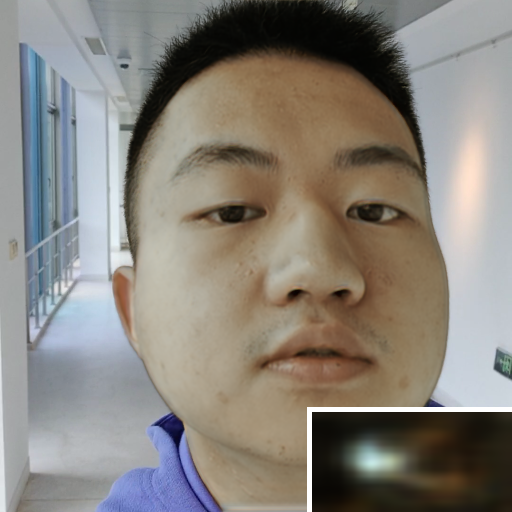}& \ig{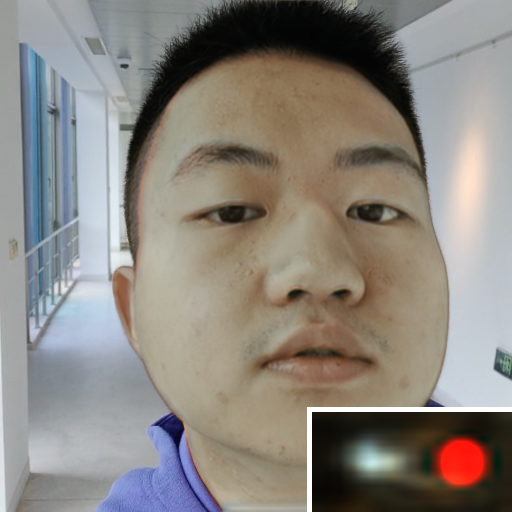}& \ig{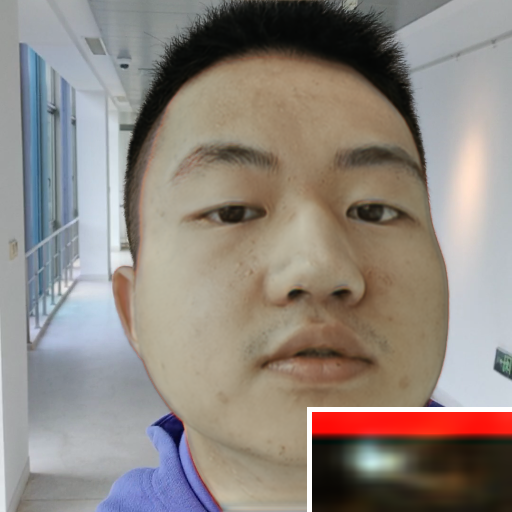}
		\\
		\rotatebox{90}{with}  & \ig{origin} & \ig{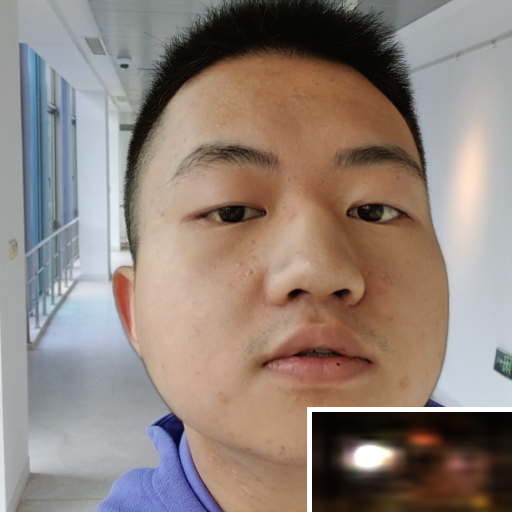}& \ig{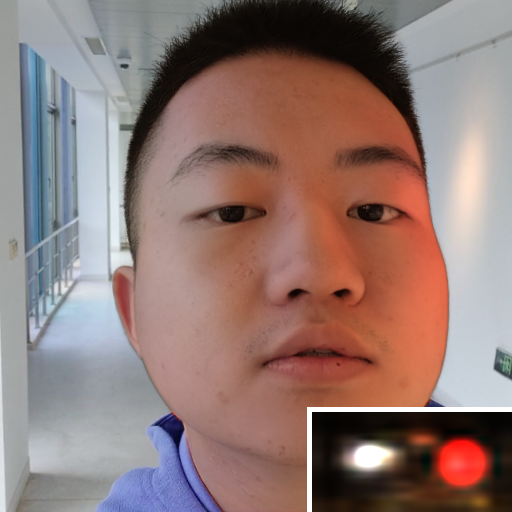}& \ig{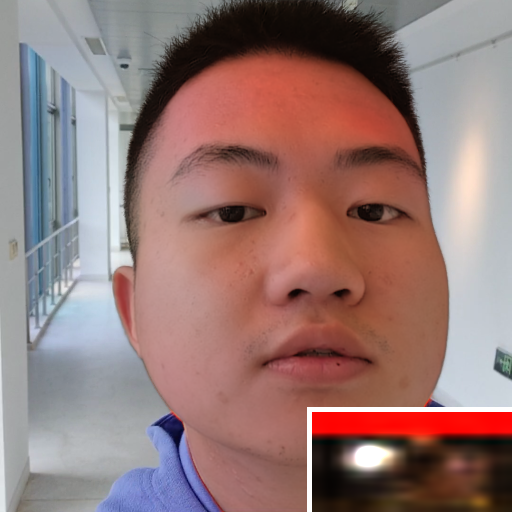}
		\\
		& Input & Recon & Modified & Modified
		\\ 
	\end{tabular}
	\caption{Evaluation of lighting sampling on real-world scenes. Our scheme enables more natural lighting manipulation, where a red light source is added into the environment lighting map. }
	\label{fig:without_sampling}
\end{figure}

\begin{figure}[t]
	\centering
	\setlength{\tabcolsep}{0.5pt}
	\newcommand{\figsize}{1.9cm}
	\newcommand{\ig}[1]{\includegraphics[width=\figsize,height=\figsize]{./fig/evaluation/sampling/#1.png}}
	
	\begin{tabular}{cccccccc}
		\ig{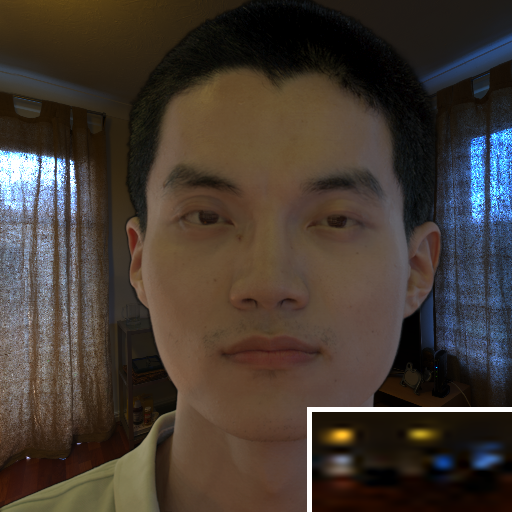} & \ig{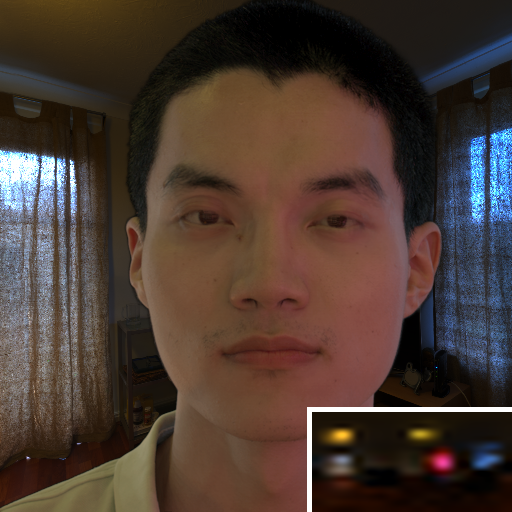}& \ig{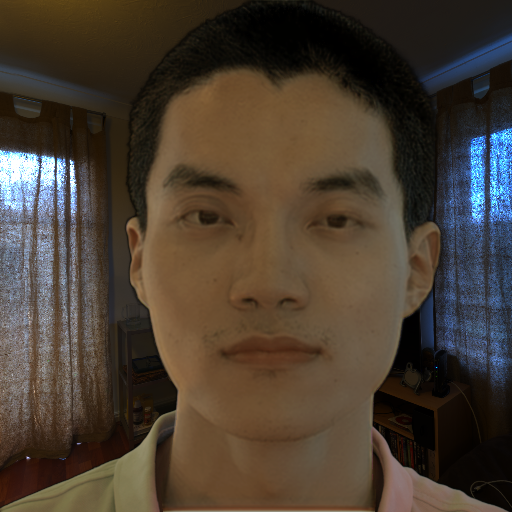}& \ig{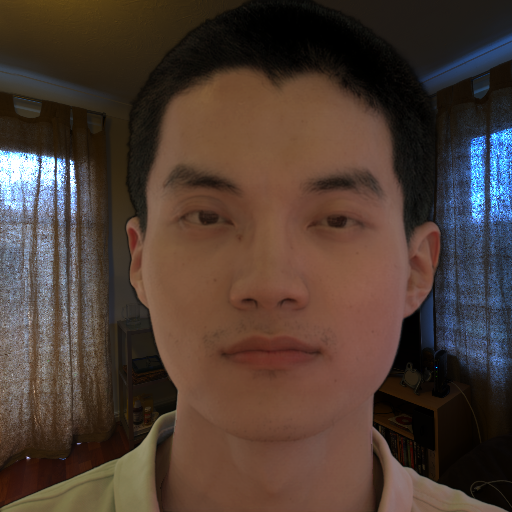}
		\\
		Input & GT editing & w/o sampling & with sampling
	\end{tabular}
	
	\includegraphics[width=\columnwidth]{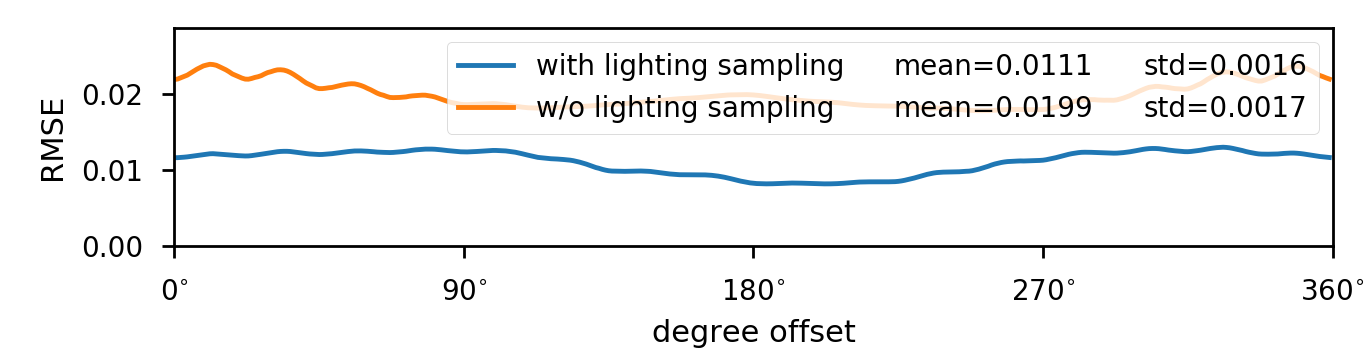}
	\caption{Evaluation of lighting sampling on synthetic scenes where a red light is added from different directions. Top: the qualitative relit examples. Down: the corresponding error curve.}
	\label{fig:without_sampling2} 
	\vspace{-10pt}
\end{figure}

\myparagraph{Lighting Sampling}
We further evaluate our light sampling strategy. 
Let \textbf{w/o~sampling} denote our variation only using the discrete environment lighting maps during training.
As shown in Fig.~\ref{fig:without_sampling} and Fig.~\ref{fig:without_sampling2}, our scheme models the illumination consistency and mutation, enabling a more natural portrait light-editing and relighting.

We further perform thorough quantitative analysis of the individual components of our approach using our testing set with ground truth.
As shown in Tab.~\ref{table:evaluation}, our full pipeline consistently outperforms other variations.
This not only highlights the contribution of each algorithmic component but also illustrates that our approach enables consistent video portrait relighting.

\section {Discussion and Conclusion} 

\myparagraph{Limitations.}
As a trial to explore real-time and consistent video portrait relighting under dynamic illuminations, our approach still owns limitations as follows.
First, our approach cannot handle the extreme illumination changes not seen in training, like suddenly turn on/off all the lights.
Besides, the generated relit video results lose the facial details partially due to the light-weight encoder-decoder architecture.
We plan to utilize a parametric face model to recover the facial details explicitly. 
Our current approach is also limited to headshots only. 
It's promising to include clothes and garment material analysis for full-body portrait relighting.
It's also interesting to enhance the OLAT dataset with a generative model to handle high-frequency lighting.

\myparagraph{Conclusion.}
We have presented a novel scheme for real-time, high-quality and consist video portrait relighting under dynamic illuminations from monocular RGB stream and a new dynamic OLAT dataset.
Our hybrid disentangles scheme with a multi-task and adversarial training strategy models the semantic consistency efficiently and generates realistic relit results.
Our temporal modeling scheme encodes the flow-based supervision for temporally consistent relighting, while our light sampling strategy enhances the illumination consistency for natural portrait lighting manipulation.
Extensive results demonstrate the effectiveness of our approach for consistent video portrait relighting in dynamic real-world scenarios.
We believe that our approach is a critical step for portrait lighting analysis, with many potential applications in user-facing photography, VR/AR visual effects or immersive telepresence.

{\small
\bibliographystyle{ieee_fullname}
\bibliography{egbib}
}

\end{document}